\documentclass{bmvc2k}
\usepackage{booktabs}
\usepackage{amsmath,amssymb}
\usepackage{mathtools}
\usepackage{dsfont}
\usepackage{multirow}
\usepackage{graphicx}
\usepackage{pgfplots}
\usepackage{tikz}

\def\red#1{\textcolor[rgb]{1,0,0}{#1}}

\newcommand{\keypoint}[1]{\vspace{0.1cm}\noindent\textbf{#1}}
\definecolor{deepGreen}{RGB}{0,153,0}
\usepackage{hyperref}

\title{Cross-Modal Hierarchical Modelling for Fine-Grained Sketch Based Image Retrieval}

\addauthor{Aneeshan Sain}{a.sain@surrey.ac.uk}{1,2}
\addauthor{Ayan Kumar Bhunia}{a.bhunia@surrey.ac.uk}{1}
\addauthor{Yongxin Yang}{yongxin.yang@surrey.ac.uk}{1,2}
\addauthor{Tao Xiang}{t.xiang@surrey.ac.uk}{1,2}
\addauthor{Yi-Zhe Song}{y.song@surrey.ac.uk}{1,2}

\addinstitution{
 SketchX, CVSSP\\ 
 University of Surrey, UK\\
}
\addinstitution{
 iFlyTek-Surrey Joint Research Centre on Artificial Intelligence\\ 
}
  
\runninghead{Sain, Bhunia, Song}{Cross-Modal Hierarchical Modelling for FG-SBIR}

\def\etal{\emph{et al}\bmvaOneDot}

\begin{document}
\maketitle
\begin{abstract}
{
Sketch as an image search query is an ideal alternative to text in capturing the fine-grained visual details. Prior successes on fine-grained sketch-based image retrieval (FG-SBIR) have demonstrated the importance of tackling the unique traits of sketches as opposed to photos, e.g., temporal vs.~static, strokes vs.~pixels, and abstract vs.~pixel-perfect. In this paper, we study a further trait of sketches that has been overlooked to date, that is, they are hierarchical in terms of the levels of detail -- a person typically sketches up to various extents of detail to depict an object. This hierarchical structure is often visually distinct. In this paper,  we design a novel network that is capable of cultivating sketch-specific hierarchies and exploiting them to match sketch with photo at corresponding hierarchical levels. In particular, features from a sketch and a photo are enriched using cross-modal co-attention, coupled with hierarchical node fusion at every level to form a better embedding space to conduct retrieval. Experiments on common benchmarks show our method to outperform state-of-the-arts by a significant margin.
% against a typical triplet loss based siamese baseline.
}
\end{abstract}

%-------------------------------------------------------------------------
\vspace{-0.4 cm}
\section{Introduction}
\vspace{-0.1cm}
\label{sec:intro}
Great strides have been made towards cross-modal image retrieval~\cite{ nam2017dual, gu2018look}, predominantly using text as a query \cite{lee2018stacked, wang2019camp}. Sketches as an alternative query mainly focussed on {category-level retrieval} at first~\cite{wang2015sketch, collomosse2017sketching, bui2018deep, collomosse2019livesketch}. 
It was not until recently when the fine-grained characteristic of sketches has been noted, which then triggered the study on fine-grained image retrieval~\cite{yu2016sketch, song2017deep, pang2019generalising}.
% Sketches can optimally model fine-grained structural details which are inexplicable via tags [14] or words [22]. This shifted current focus from categorical SBIR [2, 8] to fine-grained SBIR (FG-SBIR) [31, 40, 54]. 
At its inception, \textit{fine-grained} sketch-based image retrieval (FG-SBIR) focussed on retrieving photos of commercial objects \cite{yu2016sketch, pang2019generalising, bhunia2020sketch}. This had very recently been extended to the problem of graphical user interface (GUI) retrieval~\cite{huang2019swire}, where GUI sketches are used to conduct fine-grained retrieval of mobile app screenshots.

%which Retrieving GUI-images from large datasets is difficult as their functional and structural details are indirectly reflected only in their pixel and meta-data. As a GUI's defining structure is inexplicable via text/tag, sketch forms an optimal query medium, thus introducing GUI-image retrieval as a recent FG-SBIR application.
%A sketch is an optimal query medium for retrieving a GUI-image, as its defining structure is quite ambiguous to describe via text/tag, 
%Fig1_newfree_level

\begin{figure}[t]
\begin{center}
  \includegraphics[width=1.0\linewidth]{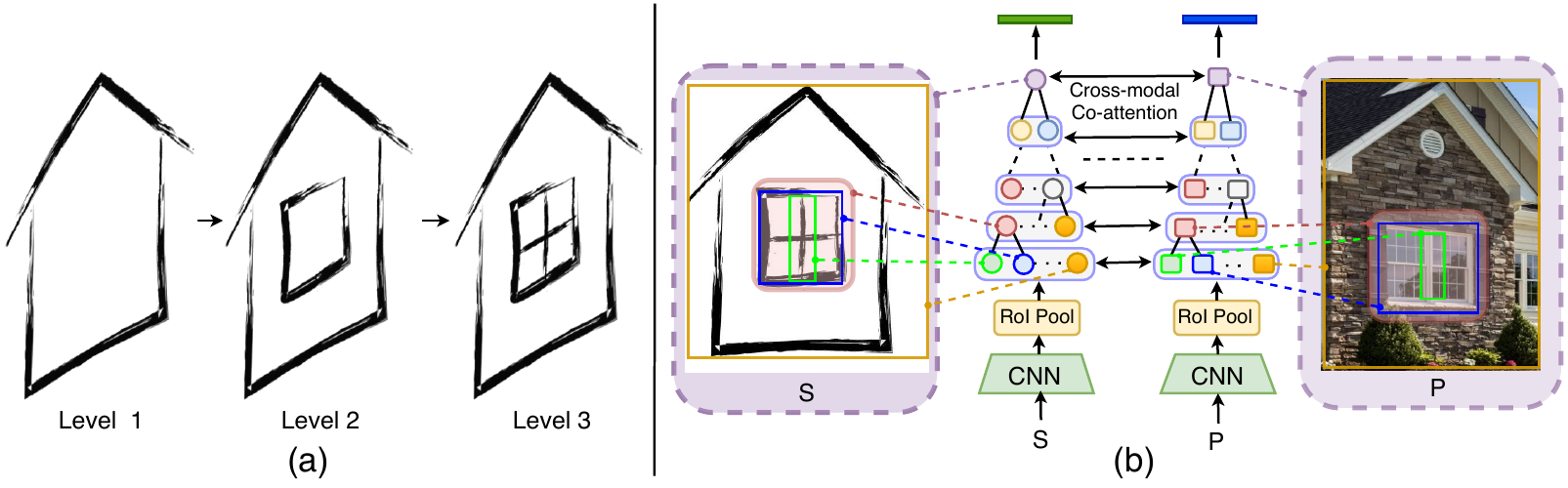}
\end{center}
\vspace{-0.75cm}
  \caption{
  {(a) shows different hierarchical levels of detail in a free-hand sketch (illustrative purpose only), while (b) illustrates our overall idea of cross-modal hierarchical modelling.}
  }
\label{fig:methods}
\vspace{-0.6cm} 
\end{figure}

It has been shown that by specifically targeting the unique traits of sketches, the sketch-photo domain gap can be effectively narrowed. Exemplar works include those that address the sequential~\cite{46008}, abstract~\cite{umar2019goal}, and stroke-wise~\cite{bhunia2020sketch} characteristics of sketches, either separately or in combination~\cite{sampaio2020sketchformer, xu2018sketchmate}. However, all previous works predominately treat sketches as a single flat structure, without recognising the inherent hierarchical structure within. Hierarchies in a sketch importantly underpin its flexibility as a search query -- when an user sketches, the \textit{extent of details being sketched} can vary from coarse to fine. 
% An object's details can vary from coarse-level to fine-grained details in a sketch.
{As Figure~\ref{fig:methods}(a) shows, a house may be drawn as a simple outline (Level 1), with further strokes denoting the window (Level 2), or with even finer strokes to depict window pattern (Level 3).}
%Considering extent of details in a hierarchical sense, a representation of that box and the combined representation of window panes and bars, should lie at a similar hierarchical level.
Understanding sketching hierarchies is crucial in handling the uncertainty brought by the varying levels of sketch details. %Learning this could therefore ensure that \textit{whichever {level of detail} an user draws to}, the model would still retain its retrieval accuracy. 
Importantly, it enables the learning of discriminative features for sketch-to-photo matching at different levels of detail. 

% Existing FG-SBIR works \cite{yu2016sketch, song2017deep} had {independently} embedded sketches and images in a common space. Although features of gallery images could be pre-computed to accelerate retrieval, it disregards cross-modal interaction. This leads to sub-optimal features as the query representation is unaware of its paired image.
% Alleviating this, Varior \etal~\cite{varior2016gated} employed a gating mechanism in person re-identification task, comparing mid-level network features. While they focus on global feature maps, Wang \etal~\cite{wang2019camp} involves local image sub-regions, capturing cross-modal correlation better.
% As similar pairwise embedding lacks usage in FG-SBIR, we introduce region-wise cross-modal feature matching between sketches and photos.

% {reshuffle text from below to make a paragraph on why modelling hierarchies for fgsbir is difficult, and why prior approaches can not be simply adapted}

% {reorganise text below to say how we did it, (i) on modelling the hierarchy, and (ii) on utilising the hierarchies for sbir}

Devising a hierarchy amongst sketching elements is non-trivial as they do not follow any predefined composition rules. Rather, sketches exhibit variable details and hierarchical compositions, due to differences in subjective perception and sketching skill.
% Beside exhibiting variable details, hierarchical composition would also vary for every sketch-photo pair, thus making human-annotated hierarchy infeasible.
% Variable extent of details coupled differing sketch-photo pair
Furthermore, in the context of sketch-photo retrieval, it is crucial for both modalities to withhold cross-modal semantic relevance at corresponding hierarchical levels.
% To handle this, cross-modal correlation at corresponding hierarchical levels is important.
In this paper, we propose a novel FG-SBIR framework that discovers the hierarchical structure within each modality, where the discovery is aided by exchanging cross-modal information at each level.
% That is, hierarchy formation in each modality, is not only dependent on its self-modal knowledge, but also on that from its matching modality. 
This cross-modal hierarchy building has two vital benefits: (i) it helps with the discovery of underlying semantic hierarchy, especially for sketches which exhibit highly sparse visual cues compared to photos, 
% e.g., having matching photo of a sketch will aid in its semantic parsing,
and (ii) it encourages the learning of a better cross-modal embedding by exchanging information across two domains at each level and propagating across different levels to form the final embedding vectors.

To establish a hierarchy in each modality, we {mimic} an agglomerative merging scheme where {the model chooses} two nodes to be merged at every level to form the next. {Albeit desirable, merging more than two nodes simultaneously would make the problem computationally intractable~\cite{choi2018learning}.
}The key question therefore comes down to \textit{choosing} which two nodes to merge. Such a choice involves a \textit{discrete} decision, thus invoking non-differentiability into the network.
%As our first contribution, we propose a straight-through Gumbel-Softmax~\cite{jang2016categorical} operation 
As our first contribution, {we model this choice via a straight-through Gumbel-Softmax~\cite{jang2016categorical} operation that approximates one-hot vectors sampled from a distribution by making them continuous ($\S$\ref{sec:hierarchy}). This helps with calculating gradients of discrete decisions thus allowing backpropagation, and making the merging scheme end-to-end trainable.}
Upon discovery, two nodes are merged over a learnable layer forming a higher order semantic.

% While, this hierarchy remains unexplored in treating sketch as a rasterised image in a CNN network, using sketch as a coordinate \cite{46008} (modelled via RNN) does not help either.

%It achieves that by {computing}
In order to encourage the learning of a better cross-modal embedding, we further constrain node merging via a cross-modal interaction module, {which computes} feature similarity across the two modalities at every hierarchy level. This is fundamentally different to prior works~\cite{yu2016sketch,song2017deep} that {independently} embed sketches and photos into a common space {without} any {cross-modal interaction}.
{More specifically, we introduce a cross-modal co-attention module that attends to salient regions in a sketch and its photo counterpart, thus ensuring mutual awareness of both modalities.
%computing and exchanging interactive information between them
% Region features from image are used as cues to attend on the sketch and vice-versa.
Information aggregated from the photo branch (based on attentive cues from the sketch branch), is integrated with the sketch branch via a gating mechanism; and vice-versa. The gating mechanism adaptively controls the fusion intensity thus filtering out negative effects of mismatched parts.}

%Note that different to prior work on modelling graphs \cite{xu2017scene,vaswani2017attention} that only learns internal relations within a single modality, we differ in modelling a sketch hierarchically cross modalities, and following a {coarse-to-fine scheme}. Moreover, graph-based methods rely on a predefined or heuristic-based adjacency matrix \cite{teney2017graph}, which is absent in sketches. 

% An off-the-shelf choice for hierarchical modelling could have been Recursive Neural Networks (RvNN), that capture linguistic composition in texts~\cite{irsoy2014deep}. RvNNs however work in a {single modality} requiring a structured input~\cite{socher2013reasoning}, thus making them unsuitable for {cross-modal feature matching}, or modelling sketches without predefined structure. 
% Consequently, to ensure a better feature representation we introduce region-wise cross-modal feature matching between sketches and photos in our work.
% Even if such methods from other related works implementing cross-modal interaction could be adapted as off-the-shelf components for FG-SBIR, the inherent hierarchy of strokes would still be ignored. 
% This upholds FG-SBIR's underlying challenge of establishing a dense correspondence using spatial information from a rasterised sketch.
%On the contrary, we aim to self-discover the hierarchy from a given sketch without any predefined constraints, additionally employing \emph{cross-modal co-attention for feature matching at every hierarchical level}.
% -- a generalization of RNNs
{In summary, our contributions are: (i) an end-to-end trainable architecture that enables the discovery of the underlying hierarchy of a sketch, (ii) a co-attention module to facilitate cross-modal hierarchy construction, and (iii) an unique perspective of utilising hierarchies for the problem of FG-SBIR.
Extensive ablative studies and evaluations against state-of-the-arts on three standard FG-SBIR and GUI retrieval datasets, show our method to outperform most existing methods by a significant margin.}

\vspace{-0.6cm}
\section{Related Work}
\label{sec:related}
\vspace{-0.35cm}

\noindent \textbf{Category-level SBIR:} Most category-level SBIR approaches fall into two categories:
(i) {handcrafted descriptors}~\cite{saavedra2015sketch, tolias2017asymmetric} which constructs global \cite{qi2015making} or local \cite{hu2013performance} photo-sketch joint representations. (ii) {deep learning methods}~\cite{yu2016sketch, song2017deep, liu2017deep},
where classical ranking losses, like contrastive or triplet loss have been used.
% Contemporary research directions include zero-shot SBIR and sketch-photo hashing.
%Unlike traditional SBIR tasks where training and testing share the same object classes \cite{cao2011edgel, collomosse2017sketching, collomosse2019livesketch, bui2018deep}, 
Other related problems such as zero-shot SBIR \cite{dey2019doodle, dutta2019semantically, liu2019semantic} and sketch-photo hashing \cite{liu2017deep, shen2018zero} 
are also studied. %aims to reduce computational cost by embedding to binary hash-codes instead of continuous vectors.

\noindent \textbf{Fine-grained SBIR:} Unlike category-level SBIR, fine-grained SBIR aims at instance-level matching. This meticulous task relies on learning the unique traits of a sketch that lie in its fine-grained details.  Being fairly new, FG-SBIR is less studied in comparison to category-level SBIR tasks.
Starting from the study of deformable-part models~\cite{li2014fine}, recent methods have learned to extract comparable features from  heterogeneous domains \cite{yu2016sketch, song2017deep, pang2017cross}.
Yu \etal \cite{yu2016sketch} proposed a deep triplet-ranking model for instance-level FG-SBIR, which was enhanced via hybrid generative-discriminative cross-domain image generation \cite{pang2017cross} and attention based techniques, in addition to advanced triplet loss formulations \cite{song2017deep}. Analogous to `zero-shot' SBIR tasks, Pang \etal \cite{pang2019generalising} worked on cross-category FG-SBIR. Recently, Huang \etal \cite{huang2019swire} applied a triplet network to retrieve GUI photos from sketch \cite{huang2019swire} and contributed the first GUI sketch dataset {SWIRE~\cite{huang2019swire}}.
These methods however, do not involve any cross-modal correlation in the context of FG-SBIR. Furthermore the notion of hierarchy has not yet been applied in this context.

\noindent \textbf{Cross-modal Interaction:}
% \{Recursive Neural Networks (RvNN), that capture linguistic composition in texts~\cite{irsoy2014deep} could have been an off-the-shelf choice for hierarchical modelling, however their operation in a {single modality} requiring a structured input~\cite{socher2013reasoning}, hinders {cross-modal feature matching}, or modelling sketches without predefined structure.
% Despite being popular in parallel literature~\cite{varior2016gated, wang2019camp}, cross-modal interaction lack usage in context of FG-SBIR.}
Existing FG-SBIR works \cite{yu2016sketch, song2017deep} \textit{independently} embed sketches and photos into a common space, disregarding cross-modal interaction. This leads to sub-optimal features as the query representation is unaware of its paired photo during training. 
Others in the non-sketch literature had otherwise successfully investigated the benefits of cross-modal interaction. Varior \etal~\cite{varior2016gated} employed a gating mechanism for person re-identification, comparing mid-level network features across both modalities. While they focus on global feature maps, Wang \etal~\cite{wang2019camp} involved local image sub-regions, capturing cross-modal correlation better. 
{
% We present ViLBERT (short for Vision-and-Language BERT), a model for learning task-agnostic joint representations of image content and natural language. We extend the popular BERT architecture to a multi-modal two-stream model, processing both visual and textual inputs in separate streams that interact through co-attentional transformer layers.
Recently, ViLBERT~\cite{lu2019vilbert} leveraged the BERT framework to interact separately processed visual and textual inputs via co-attentional transformer layers.} In this paper, we introduce cross-modal feature matching between sketches and photos, and show how it can be embedded during hierarchy construction.

%li2019graph
\noindent \textbf{Hierarchical Modelling:}
Several works have emerged modelling latent textual hierarchy using Recursive Neural Networks (RvNN) \cite{socher2013reasoning}.
Tree-RNNs~\cite{socher2011parsing}, enhanced by RNTN networks~\cite{socher2013recursive} surpassed vanilla LSTMs in synthesising sentence meaning from word-vectors. Stacking tree-structured vanilla RvNNs~\cite{irsoy2014deep} improved sentiment classification. While gated RvNNs \cite{cho-etal-2014-properties} controlled children-to-parent information flow, Neural tree indexer (NTI) \cite{munkhdalai2017neural} used soft hierarchical structures to employ Tree-LSTMs. Methods using discrete levels of contextual information~\cite{zhu2011recursive}, recursive context propagation networks~\cite{sharma2014recursive}, shape-parsing strategies~\cite{liu2015understanding} or scene-graphs~\cite{johnson2015image, xu2017scene} have aimed at encoding regions with semantic relations.
{
Graph neural networks (GNN) \cite{qi2018learning, chen2015gated} being well-suited to structured data offer viability towards implementing hierarchy.
% Albeit graph neural networks (GNN) \cite{qi2018learning, chen2015gated} are well-suited to structuring hierarchy, they inherently rely upon a predefined structure thus being not a good fit for sketches. 
DIFFPOOL~\cite{ying2018hierarchical} learns a differentiable soft cluster assignment to map nodes into clusters, which are pooled as input for the next GNN layer. 
Some works aim at capturing contextual structure \cite{teney2017graph, xu2017scene} by building a graph between scene objects and texts. Recent works like SAGPool~\cite{lee2019self} and gPool~\cite{gao2019graph} focussed on top-K node selection to formulate a subgraph for the next GNN layer. While EdgePool~\cite{diehl2019edge} reduces almost half its nodes per GNN layer by contracting its edges, EigenPool~\cite{derr2018signed} controls pooling ratio based on graph Fourier transform. Nevertheless, all such approaches usually leverage a pre-defined structure.
}
Sketches however, are devoid of such strict composition rules, but harbour an implicit hierarchy which our model learns to discover in this work.

\begin{figure*}[t]
\centering
\includegraphics[width=\linewidth]{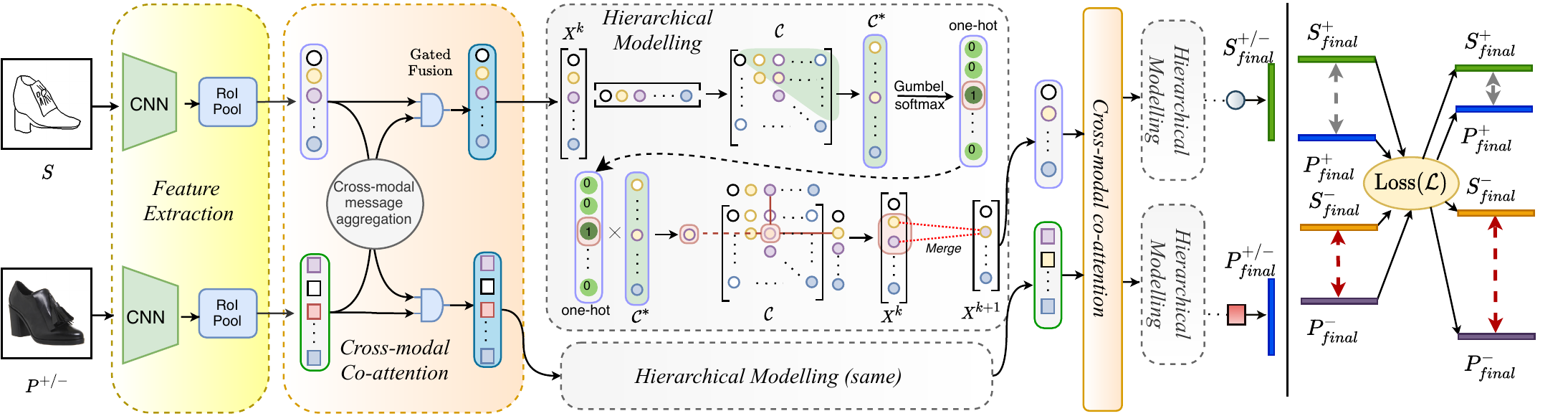}
\vspace{-0.6cm}
\caption{
{Our framework. After extracting region-wise features from a sketch (S) and a photo ($P^{+/-}$), they are enriched via cross-modal attention ($\mathcal{M}$), followed by hierarchical parsing operation ($h_\phi$). The last two steps are repeated consecutively, until a final representation from each branch ($S_{final}^{+/-}$,$P_{final}^{+/-}$) is obtained. A loss ($\mathcal{L}$) brings the matching sketch-photo pair ($S_{final}^+$,$P_{final}^+$) {closer} (grey) while {distancing} (red) the unmatched one ($S_{final}^-$,$P_{final}^-$).}
}
\label{fig:network}
\vspace{-0.4cm}
\end{figure*}

\vspace{-0.3cm}
\section{Methodology}
\label{sec:methodology}
\vspace{-0.2cm}
\noindent \textbf{Overview:} 
We aim to embed sketches and photos into a common space such that, (a) the underlying hierarchy of a sketch is modelled, and (b) feature learning in either modality is reinforced via cross-modal interaction at every hierarchical level. Formally, we learn an embedding function $F: I \rightarrow \mathbb{R}^d$ that maps a photo or rasterised sketch to a $d$ dimensional feature, which we use to retrieve against a distance metric. We employ three modules:

\noindent \textbf{Backbone Feature Extractor:} Let $f_{\theta}$ be a {backbone feature extractor} initialised from pre-trained InceptionV3~\cite{bhunia2020sketch} weights. Using input $I$, we obtain a feature map $\mathcal{F}_I = f_{\theta}(I) \in \mathbb{R}^{H\times W \times C}$ where $I$ denotes a sketch $S$ or RGB photo $P$, and $H$, $W$, $C$ signifies height, width and number of channels respectively. To extract region-wise features, we treat every sketch-stroke as an individual region, and calculate bounding boxes enclosing every stroke. Meanwhile, we use the unsupervised selective search algorithm~\cite{uijlings2013selective} to discover sub-regions from $P$. 
Let $S$ and $P$ have $N_{S}$ strokes and $N_{P}$ discovered sub-regions with bounding box sets of $B_{S} \in \{b_{1}^s, b_{2}^s, ..., b^s_{N_{S}}\}$ and $B_{P} \in \{b^p_{1}, b^p_{2}, ..., b^p_{N_{P}}\}$ respectively.
We perform ROI pooling on feature maps $\mathcal{F}_{S}$ and $\mathcal{F}_{P}$ using $B_{S}$ and $B_{P}$, to obtain sets of region-wise latent feature vectors ${S_{r}} = \{s_{1}, s_{2}, ..., s_{N_{S}}\}$ and ${P_{r}} = \{p_{1}, p_{2}, ..., p_{N_{P}}\}$; where $S_r \in \mathbb{R}^{N_{S}\times  C}$ , $P_r \in \mathbb{R}^{N_{P}\times C}$ (as matrices).
A linear projection layer then reduces each vector of ${P_{r}}$ and ${S_{r}}$ to a $d$-dimensional space.

% \noindent \textbf{Hierarchical Parsing Module:} This module (denoted as $h_{\phi}$) takes $S_r$ or $P_r$ as input and constructs a hierarchy of multiple levels. At the hierarchical level $k$, $h_{\phi}$ outputs  ${X^k} = \{ x_{j}\}_{j=1}^{N^k}, x_{j} \in \mathbb{R}^d $ consisting of a set of $N^k$, $d$-dimensional vectors (nodes). It does so by starting with $S_r$ or $P_r$ and progressively identifies  and merges two nodes at a time to form a higher order semantic representing the next hierarchical level:  ${X}^{k+1} = \{ x_{j}\}_{j=1}^{N^{k+1}}, \; x_{j} \in \mathbb{R}^d, \; N^{k+1}=N^k-1$. This node-merging process continues until a final single $d$-dimensional vector ($\mathbb{R}^d$) remains per branch (see Sec.~\ref{sec:hierarchy} for details and Figure~\ref{fig:network} for an illustration). % -- by Tony.
\noindent \textbf{Hierarchical Parsing Module:} This module (denoted as $h_{\phi}$) takes $S_r$ or $P_r$ as input and constructs a hierarchy of multiple levels. At any hierarchical level $k$, $h_{\phi}$ {inputs}  ${X^k} = \{ x_{j}\}_{j=1}^{N^k}, x_{j} \in \mathbb{R}^d $ consisting of a set of $N^k$, $d$-dimensional vectors (nodes). It starts with $S_r$ or $P_r$ and progressively identifies  and merges two nodes at a time to form a higher order semantic, representing the next hierarchical level:  ${X}^{k+1} = \{ x_{j}\}_{j=1}^{N^{k+1}}, \; x_{j} \in \mathbb{R}^d, \; N^{k+1}=N^k-1$. This node-merging process continues until a final single $d$-dimensional vector ($\mathbb{R}^d$) remains per branch (see Sec.~\ref{sec:hierarchy} for details and Figure~\ref{fig:network} for an illustration).

\noindent \textbf{Cross-modal Co-attention:} While $h_\phi$ is able to discover the underlying hierarchy of a sketch, we additionally stress on cross-modal co-attention between sketch and photo branches, at every hierarchical level. More specifically, we design a \textit{cross-modal co-attention module} $\mathcal{M}$ which takes both ${P_{r}^k}$ and ${S_{r}^k}$ at the $k^{th}$ hierarchical level, aggregates sketch branch information onto the photo branch and vice-versa, and finally outputs the modified vectors ${\tilde{P}_{r}^{k}}$ and ${\tilde{S}_{r}^{k}}$. 
Overall, $\mathcal{M}$ enforces cross-modal interaction on the sets of vectors from photo branch ${P_{r}^k}$ and sketch branch ${S_{r}^k}$ having region-wise features  at the $k^{th}$ hierarchical level (see Sec~\ref{sec:cross-modal}).  
%After information exchange across modalities using this co-attention module,
{Following such cross-modal information exchange,} $h_\phi$ is applied to merge two nodes in each branch at each level.
{These two steps occur consecutively at every hierarchical level, until one node remains in each branch.}
% Following sections describe our Hierarchical Modelling (\S~\ref{sec:hierarchy}) and Cross-modal Co-attention modules (\S~\ref{sec:cross-modal}) in details. 

\vspace{-0.3cm}
\subsection{Hierarchical Modelling}
\label{sec:hierarchy}
\vspace{-0.25cm}
Given a set of region-wise features $X^k = \{x_i\}^{N^k}_{i=1}, \; x_i \in \mathbb{R}^d$ at the $k^{th}$ hierarchical level, $h_\phi$ aims to measure the affinity between all possible unique pairs of nodes and merge the two having the highest validity score to form a higher order semantic. 
We approach this by projecting the node features to a low $d_h$-dimensional space and formulating an intra-regional compatibility matrix $\mathcal{C} = (X^k \cdot \mathbf{W}^\mathcal{C}_\phi)\cdot( X^k \cdot \mathbf{W}^\mathcal{C}_\phi)^\mathrm{T}, \; \mathcal{C} \in \mathbb{R}^{N^k \times N^k}$
where $\mathbf{W}^\mathcal{C}_\phi \in \mathbb{R}^{d\times d_h}$ is a projection matrix. $\mathcal{C}_{i,j}$ represents the validity score of each pair of nodes.
Since $\mathcal{C}$ is symmetric, we consider only its upper triangular elements, excluding the principle diagonal: 
$\mathcal{C}^* =$ \texttt{flatten}(\texttt{UpTri}($\mathcal{C}$)), $\mathcal{C}^* \in \mathbb{R}^{H^k}$ 
where $H^k = \frac{N^k(N^k-1)}{2}$, \texttt{UpTri}() extracts upper-triangular elements of a matrix and \texttt{flatten}() compiles them into a vector. 
The pair having the highest compatibility score shall intuitively suggest the highest probability of merging. 

%Overcoming non-differentiability
Using \textit{argmax} here is non-differentiable and might naively need Monte Carlo estimates with a REINFORCE-type algorithm \cite{williams1992simple}, which would typically suffer from high variance~\cite{jang2016categorical}. We thus apply a low-variance gradient estimate via Gumbel-softmax
\linebreak re-parameterisation trick and Straight-Through (ST) gradient estimator~\cite{jang2016categorical} to $\mathcal{C}^*$. Gumble-softmax approximates one-hot vectors sampled from a distribution by introducing a \textit{Gumbel noise}~\cite{jang2016categorical} in every sampling step. This replaces the discontinuous \textit{arg-max} function by a differentiable softmax function.
% In essence, given a $H^k$-dimensional categorical distribution across every node-pair, with a set of unnormalised probabilities, $\mathcal{C}^* =  (c_1,c_2,...c_{H^k})$, a sample \textbf{q}$= (q_1,q_2,...q_{H^k})$ from Gumbel-softmax distribution can be drawn as:
Given a $H^k$-dimensional categorical distribution across every node-pair, with $\mathcal{C}^* =  (c_1,c_2,...c_{H^k})$, where $c_i = \log(\pi_i)$ and $\pi_i$ is an unnormalised log probability, we draw a sample \textbf{q}$= (q_1,q_2,...q_{H^k})$ from Gumbel-softmax distribution as:  
\vspace{-0.1cm}
\begin{equation}
\label{equ:gumbel}
\begin{aligned}
q_i = \frac{\exp{((\log(\pi_i) + g_i)/\tau})}
{\sum^{H^k}_{j=1}\exp{((\log(\pi_j)+g_j)/\tau})},
\end{aligned}
\vspace{-0.1cm}
\end{equation}
where, $g_i$ represents \textit{Gumbel-noise} and $\tau$ is the temperature parameter~\cite{chorowski2015attention}. As $\tau$ tends to 0, \textbf{q} resembles a one-hot sample. 
In the forward pass it discretises a continuous probability vector \textbf{q} sampled from Gumble-Softmax distribution into a one-hot vector $\mathbf{q}^{ST} = (q^{ST}_1,q^{ST}_2,...q^{ST}_{H^k})$ where $\mathbf{q}^{ST}_i = \mathds{1}_{[i \;= \; \mathrm{argmax}_j(q_j)]}$.
During backward pass, it uses the continuous \textbf{q}, thus allowing backpropagation.
If $\mathcal{C}_{a,b}$ is the element for which $q^{ST}_i$ equals 1, we fuse nodes $x_a$ and $x_b$ as:
%One out of $H^k$ node-pairs, is fused based on one-hot sampling from $\mathbf{q}^{ST}$ as : 
$\hat{x}_{a,b} = \texttt{ReLU}(\mathbf{W}^F_\phi \cdot [x_a, \; x_b]), \; \mathbf{W}^F_\phi \in \mathbb{R}^{d\times 2d}$
%[$x_a,x_b$] denotes vector concatenation;
leading to the next hierarchical level output: 
\vspace{-0.1cm}
\begin{equation}
\vspace{-0.1cm}
\label{equ:updation}
\begin{aligned}
{X}^{k+1} := {X}^{k} - \{x_a, x_b\} + \{\hat{x}_{a,b}\} \;\; ; \qquad x_a,x_b \in X^k.
\end{aligned}
\vspace{-0.1cm}
\end{equation}
Note that our formulation considers fusion of distant nodes in addition to adjacent ones~\cite{choi2018learning}.
% Importantly, the merging scheme is completely data-driven and learned jointly with the rest of the FG-SBIR model. - yzs
{Most importantly, the merging scheme does not require low-level nodes to be merged with low-level ones only -- this fusion is guided entirely by the loss (Equation~\ref{loss_func}) alone, with no explicit control. This makes the merging scheme completely data-driven and is thus learned jointly with the rest of the FG-SBIR model.}
Hierarchical modelling is executed in parallel in both sketch and photo branches with shared weights, so that fusion in one branch is aware of that in the other, thus ensuring a better feature matching.
% by forming a better higher order semantic sense.
Formally, $P^{k+1}_r \leftarrow h_\phi(P^k_r)$, $S^{k+1}_r \leftarrow h_\phi(S^k_r)$ and $|P^{k+1}_r| = |P^{k}_r|-1$, $|S^{k+1}_r| = |S^{k}_r|-1$, where $|\cdot|$ denotes cardinality of a set. 
In case of unequal nodes in two branches, if one reduces to $\mathbb{R}^d$ first, the other continues with hierarchical fusion until that reduces to $\mathbb{R}^d$ as well, with no new fusion on the former branch. 
Thus we have $|S^{k_{final}}_r|,\;|P^{k_{final}}_r| =1$, $k_{final}$ being the final hierarchical level.

\vspace{-0.3cm}
\subsection{Cross-modal Co-attention}
\label{sec:cross-modal}
\vspace{-0.2cm}

Given a set of region-wise latent vectors $S^k_r \in \mathbb{R}^{N_s^k\times d}$ and $P^k_r \in \mathbb{R}^{N_p^k\times d}$ from photo and sketch branches at the $k^{th}$ hierarchical level respectively, we aim to enrich $S^k_r$ and $P^k_r$ by passing fine-grained information between them. Formally: $[\tilde{S}^k_r,\tilde{P}^k_r] \leftarrow \mathcal{M}(S^k_r,P^k_r)$. Here `$k$' is dropped for notational brevity.
Photo branch features ($P_r$) are aggregated w.r.t every vector from sketch feature set $S_r$. Finally the aggregated features ($P^S_r$) are combined with the sketch branch updating it as, $\tilde{S_r} =$\texttt{fused}$(S_r, \; P^S_r)$. Similarly we obtain $\tilde{P_r} =$\texttt{fused}$(P_r,\; S^P_r)$.

Towards this goal we calculate a stroke-region affinity matrix 
$\mathbf{A} = (S_r \cdot \mathbf{W}^S_\psi )\cdot(P_r \cdot \mathbf{W}^P_\psi)^\mathrm{T}$,  $\mathbf{A} \in \mathbb{R}^{N_S\times N_P}$, where every element $\mathbf{A}_{i,j}$ represents the affinity between $i^{th}$ sketch-stroke and $j^{th}$ image-region respectively. $\{\mathbf{W}^P_\psi$, $\mathbf{W}^S_\psi\} \in \mathbb{R}^{d\times d_h}$ are corresponding projection matrices. 
Attending to every stroke feature of $S_r$ with respect to every feature from $P_r$, $\mathbf{A}$ is normalised across sketch-dimension, providing a sketch-specific stroke-region affinity matrix 
$\mathbf{A}^*_S = $\texttt{softmax}$({\mathbf{A}^\mathrm{T}}/{\sqrt{d_h}})$.
Now we accumulate all sketch branch features $S_r$ w.r.t. each photo-region on $\mathbf{A}^*_S$ to finally obtain $S^P_r = \mathbf{A}^*_S \cdot S_r, \; S^P_r \in \mathbb{R} ^{N_P\times d}$,  via dot-product based feature aggregation.
On similar notion, with an attention on $P_r$ w.r.t features from $S_r$ we obtain,  $\mathbf{A}^*_P = $ \texttt{softmax}$({\mathbf{A}}/{\sqrt{d_h}})$ and consequently $P^S_r = \mathbf{A}^*_P \cdot P_r,\; P^S_r \in \mathbb{R} ^{N_S\times d}$.
$P^S_r$ is the aggregated feature from photo branch to be fused with the sketch branch, and vice-versa for $S^P_r$.

A sketch query is not only compared with its positive photo pair, but also with other negative photos which are to be filtered out during retrieval. We thus need to suppress passing of unrelated information, as well as adjust the extent of fusing aggregated features from other modality with the original one.
Consequently, we design a learnable gate, $\mathcal{G}^S = $\texttt{sigmoid}
$([S_r, P^S_r]\cdot \mathbf{W}^S_\mathcal{G})$, $\mathbf{W}^S_\mathcal{G} \in \mathbb{R}^{2d\times d}$ where every element is normalised between 0 (no fusion) and 1 (complete fusion), adaptively controlling the extent of fusion.
Multiplying $\mathcal{G}^S \in \mathbb{R}^{N_S\times d}$, with a combination of $S_r $ and $ P^S_r$, signifies that the greater the \textit{correlation} of a region with a stroke, the further shall be the impact of fusion.
Finally, to preserve the original data of region features which should not have been fused intensively, the original features are further added to the fused ones over a residual connection, giving $\tilde{S}_r$ as,
\vspace{-0.2cm}
\begin{equation}
\label{equ:finalhat}
\begin{aligned}
\tilde{S}_r = \mathcal{Z}_S(\mathcal{G}^S \odot (S_r \oplus P^S_r)) \oplus S_r \; ,
\quad \tilde{S}_r \in \mathbb{R}^{N_S\times d},
\end{aligned}
\end{equation}
where, $\mathcal{Z}_S$ is a transformation having a ReLU activation on a linear layer; $\odot$ denotes Hadamard product and $\oplus$ is element-wise summation.
%On a similar notion, $S^P_r$ and $P_r^S$ can be used symmetrically to obtain $\tilde{P}_r$:
Similarly we get $\tilde{P}_r$ from $S^P_r$ and $P_r^S$ as:
\vspace{-0.2cm}
\begin{equation}
\label{equ:finalhat2}
\begin{aligned}
\mathcal{G}^P = \texttt{sigmoid}([P_r, \; S^P_r]\cdot \mathbf{W}^P_\mathcal{G}) \;  ; \quad
\tilde{P}_r = \mathcal{Z}_{P}(\mathcal{G}^P \odot (P_r \oplus S^P_r)) \oplus P_r \; ,
\quad \tilde{P}_r \in \mathbb{R}^{N_P\times d}.
\end{aligned}
\end{equation}

\vspace{-0.4cm}
\subsection{Learning Objective}
\label{sec:learning}
Taking independent embedding of a sketch ($S$) as an anchor, Triplet loss~\cite{yu2016sketch} aims to minimise its distance from a true-match photo ($P^+$) while maximising that from a non-matching photo ($P^-$) in a joint embedding space. We however have two representations of one sketch due to cross-modal pairwise embedding w.r.t photos namely $S_{final}^+$ (paired with $P_{final}^+$) and $S_{final}^-$ (paired with $P_{final}^-$).
On similar motivation, our loss aims at decreasing distance between $S_{final}^+$ and  $P_{final}^+$, while increasing it between $S_{final}^-$ and  $P_{final}^-$, as:
\vspace{-0.2cm}
\begin{equation}
\label{loss_func}
    \begin{aligned}
    \mathcal{L}(S_{final}^+,S_{final}^-,P_{final}^+,P_{final}^-) =
    max\{0,\; \Delta + \mathcal{D}(S_{final}^+,P_{final}^+) - \mathcal{D}(S_{final}^-,P_{final}^-)\}
    \end{aligned}
\end{equation}
% \vspace{-0.2cm}
%
where, $\mathcal{D}(a,b) = || a - b ||^2$ and $\Delta$ is the margin value. 
This loss trains our network in an end-to-end manner, guiding to learn the hierarchical representation of sketch for better retrieval.

\vspace{-0.4cm}
\section{Experiments}
\label{sec:expt}
\vspace{-0.3cm}
\subsection{Experimental Settings}
\label{sec:settings}
\noindent\textbf{Datasets:}\quad
We evaluate our method on FG-SBIR specific datasets of QMUL-Chair-V2~\cite{song2018learning} and QMUL-Shoe-V2~\cite{pang2019generalising,song2018learning}, along with SWIRE~\cite{huang2019swire} dataset which was curated using GUI examples from the RICO~\cite{deka2017rico} dataset.
Out of 6,730 sketches and 2,000 photos in QMUL-Shoe-V2, 6,051 sketches and 1,800 photos were used for training, while the rest for testing.
A training/testing split of 1,275/725 sketches and 300/100 photos respectively, has been used for QMUL-Chair-V2 dataset.
RICO~\cite{deka2017rico} contains 72,000 examples of graphical user interfaces (GUI) from 9,700 free Android apps, out of which Huang \etal~\cite{huang2019swire} used a subset of 167 apps to curate 3,802 sketches of 2,201 GUI examples.
For a fairer evaluation, we do not repeat interfaces from the same apps between training and testing sets. 

\vspace{+0.1cm}
\noindent\textbf{Implementation Details:}
ImageNet pre-trained InceptionV3 network (excluding auxiliary branch) is used as a backbone feature extractor on $299\times299$ resized images. Based on selective search~\cite{uijlings2013selective}, top 16 regions are  considered {empirically} for photo branch. As coordinate information is absent in sketches from SWIRE~\cite{huang2019swire}, individual connected components (treated as strokes) are used, to obtain their bounding box information. For QMUL datasets, bounding boxes are calculated from available coordinate stroke information.
Our model is implemented in PyTorch, {taking around 2.3~mil. parameters which is roughly 5\% more than a Siamese network baseline model (\S4.2 B-Siamese)}. It is trained with Adam optimiser, using learning rate of 0.0001, batch size of 16, a margin ($\Delta$) of 0.5, temperature ($\tau$) of 1.0 and embedding size $d=512$, for upto 200 epochs on a TitanX 12-GB GPU. 
Performance is evaluated using percentage of sketches having true-match photos appearing in the top-1 (acc.@1) and top-10 (acc.@10) lists. 

\vspace{-0.2cm}
\subsection{Competitors}
\vspace{-0.2cm}
\label{sec:competitors}
 We evaluate our model against three groups of competitors:
\textit{(i)} Evaluation against existing \textit{state-of-the-arts} (SoA): 
\textbf{Triplet-SN}~\cite{yu2016sketch} uses Sketch-a-Net~\cite{yu2016sketchAnet} as a baseline feature extractor trained using triplet ranking loss. \textbf{Triplet-Attn-SN}~\cite{song2017deep} further extended {Triplet-SN} employing spatial attention along with higher order ranking loss.
\textbf{SWIRE}~\cite{huang2019swire} uses basic VGG-A network trained with typical triplet loss for the recently introduced GUI retrieval.
\textit{(ii)} As paired-embedding is ignored in SoAs, shadowing cross-modal retrieval literature \cite{varior2016gated,hwang2012learning, gong2014multi} we design a few \underline{B}aselines (B) \textit{employing paired-embeddings}:
\textbf{B-Siamese} is a naive baseline built similar to TripletSN but replaces Sketch-a-Net with more advanced InceptionV3 as backbone feature extractor.
\textbf{B-Gated-Siamese} involves paired embedding by employing a \textit{matching gate}~\cite{varior2016gated} between spatial feature-maps from intermediate layers of photo and sketch networks.
\textbf{B-Localised-Coattn} harbours paired embeddings by employing co-attention between local photo-sketch sub-regions~\cite{wang2019camp}, without any hierarchical modelling.
\textbf{B-Graph-Hierarchy} models a graph-based method. While typical graph convolutional networks (GCNs) are inherently flat with no hierarchical mechanism, we employ DIFFPOOL~\cite{ying2018hierarchical} on local region-features from each branch along with cross-modal co-attention, thus reducing the number of nodes every time by 1, until a final d-dimensional vector is obtained. 
\textit{(iii)} In context of FG-SBIR, \textit{we verify the potential of recent sketch-embedding techniques} under category-level SBIR. 
\textbf{SketchBERT-Variant} uses a transformer encoder as a sketch feature extractor on the five-point sketch coordinate representation inspired from a recent sketch classification/retrieval work~\cite{sketchbert2020}. \textbf{SketchFormer-Variant}~\cite{sampaio2020sketchformer} focusses on jointly embedding coordinate (via Transformer) and raster image ($f_\theta$) representations, by concatenating and passing them via a two layer MLP, as a sketch query.
Similar sketch-embedding topology has been recently used in SketchMate~\cite{xu2018sketchmate} and LiveSketch~\cite{collomosse2019livesketch}, where RNN embeds coordinate representation. For both these variants, {the} transformer architecture is designed following SketchBERT. We take the input via a normal linear layer and final feature is max-pooled across time, for classification.

\begin{table}[t]
\footnotesize
\centering
\caption{Quantitative comparisons against various methods. }
\label{tab:quantitative}
\resizebox{1.0\columnwidth}{!}{
\begin{tabular}{cccccccc}
\specialrule{1.pt}{1pt}{1pt}
\multicolumn{2}{c}{\multirow{2}{*}{Methods}} & \multicolumn{2}{c}{Chair-V2} & \multicolumn{2}{c}{Shoe-V2} & \multicolumn{2}{c}{SWIRE} \\ \cline{3-8} % \cline{4-4.5} \cline{6-7}  
                            &    & acc.@1 & acc.@10 & acc.@1 & acc.@10 & acc.@1 & acc.@10  \\ %\hline
\specialrule{1.pt}{1pt}{1pt}
\multirow{3}{*}{State-of-the-arts} & Triplet-SN~\cite{yu2016sketch} & 45.65  & 84.24  & 28.71  & 71.56  & -  & -  \\
& Triplet-Attn-SN~\cite{song2017deep}  & 56.54  & 88.15  & 31.74  & 74.78  & -  & -  \\
& SWIRE \cite{huang2019swire}         & -  & -  & -  & -  & 15.90  & 60.90  \\
\hline
\multirow{4}{*}{Baselines }& B-Siamese   & 49.54  & 85.98  & 30.96  & 72.54  & 54.21  & 82.15  \\
& B-Gated-Siamese         & 53.08  & 86.34  & 32.65  & 74.24  & 62.12  & 85.65  \\
& B-Localised-Coattn        & 55.24  & 88.21  & 33.21  & 77.83  & 65.48  & 88.65  \\
& B-Graph-Hierarchy          & 58.22  & 89.97  & 34.05  & 79.54  & 66.18  & 89.32  \\
\hline
\multirow{2}{*}{Others} & SketchBERT-Variant    & 13.54 & 54.78 & 8.15 & 48.23 & - & - \\
&  SketchFormer-Variant    & 32.54 & 84.82 & 26.21 & 65.34 & - & - \\
\hline
& Proposed        & \bf 62.45 & \bf 90.71 &  \bf 36.27 & \bf  80.65 & \bf 67.23 & \bf 90.11
\\ \hline
\end{tabular}%
}
\vspace{-0.3cm}
\end{table}

\vspace{-0.3cm}
\subsection{Performance Analysis}
\label{sec:results}
\vspace{-0.15cm}
Comparative performance results are shown in Table~\ref{tab:quantitative}.
\textbf{(i)} The inferior results of \textit{Triplet-SN} and \textit{Triplet-Attn-SN} are partially due to their apparently weaker backbone feature extractor of Sketch-A-Net. As for \textit{SWIRE}~\cite{huang2019swire} we suspect this is due to its inefficient training strategy. Using a stronger InceptionV3 backbone in \textit{B-Siamese} does not add much either as it uses independent sketch-photo embeddings for retrieval.
\textbf{(ii)} \textit{B-Gated-Siamese} boosts scores as it introduces cross-modal feature matching at \textit{spatial feature level}, thus creating more robust feature representation for retrieval while \textit{B-Localised-Coattn} increases it further owing to \textit{localised region-wise co-attention} based feature aggregation, from both modalities. 
\textbf{(iii)} \textit{B-Graph-Hierarchy} represents the strongest alternative to our method, however it performs consistently worse than ours. {This is because in our model, at every hierarchical level, only the two nodes selected to be merged are updated, while all others remain the same. This importantly preserves the integrity of the untouched nodes, so that they can better take part in {cross-modal interaction} at higher levels. On the contrary for graph-based methods, all nodes are updated via information passing at every level, which dilutes node features for consequent {cross-modal interaction.}
Furthermore, graph-based approaches dictate the provision of a predefined adjacency matrix. As no such predefined rules are available for our problem, we simply assume a heuristic of all nodes being fully connected. Our model on the other hand can discover the hierarchical structure from the node features themselves, without the need for such adjacency matrices.}
\textbf{(iv)} Both \textit{Sketchformer-Variant} and \textit{SketchBERT-Variant} use independent embedding of sketches and photos, exploring the coordinate information from a sketch using recent Transformer based frameworks. As GUI sketches do not contain temporal stroke information, this baseline was not performed on them. Despite their recent success in category-level SBIR \cite{xu2018sketchmate, sampaio2020sketchformer}, results on QMUL datasets show a drop in performance even after collating the rasterised sketch-image with coordinate information in \textit{Sketchformer-Variant}.
This implies that coordinate representation of a sketch is not ideally suited for the {cross-modal FG-SBIR} tasks.

%{figs/Results_GUI.pdf}
% \vspace{-0.7cm}
\begin{figure*}[t]
\vspace{-0.2cm}
\centering
\includegraphics[width=\linewidth]{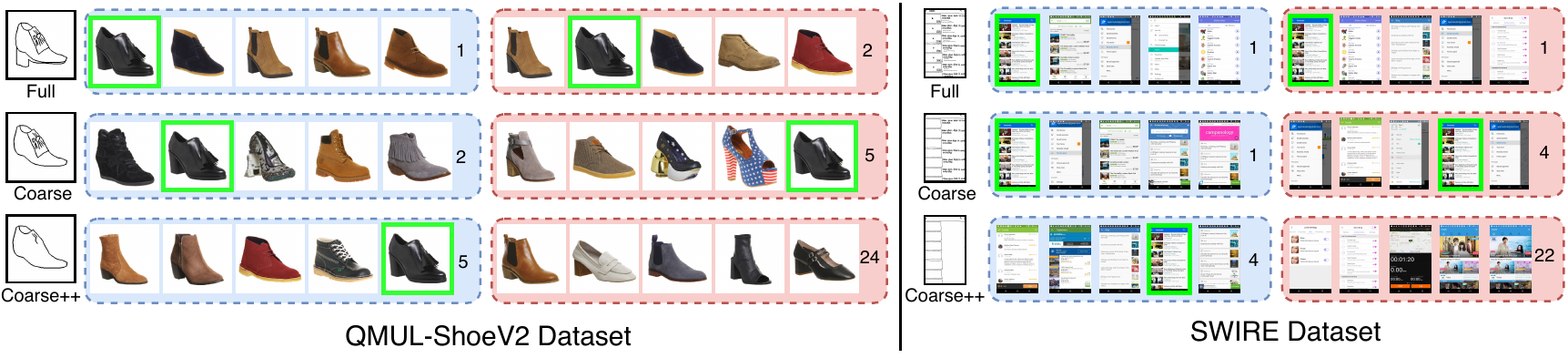}
\vspace{-0.55cm}
\caption{Our method's (blue) efficiency over \textit{B-Siamese} (red) at varying extent (\textit{coarse, coarse++}) of sketch \textit{details} is shown (\S\ref{sec:ablation}). Numbers denote rank of the matching photo.}
\label{fig: retrieval}
\vspace{-0.4cm}
\end{figure*}

\vspace{-0.3cm}
\subsection{Ablation Study}
\label{sec:ablation}
\vspace{-0.25cm}
\keypoint{Is hierarchy useful for FG-SBIR?}
To answer this we focus on GUIs~\cite{huang2019swire}, since they follow a more explicit hierarchy compared to free-hand sketches. Design elements in GUIs and their sketches exhibit a hierarchical relationship defined by containment, e.g., larger rectangular boxes representing window panes encompassing smaller ones symbolising buttons. This hierarchical information is {distinctly} defined in the meta-data released with the RICO~\cite{deka2017rico} dataset (an example shown in Figure.~\red{4}).
%In a GUI sketch we can define this layout-order using a weak assumption that, a larger connected component in a GUI sketch, encompasses a smaller one inside it.
This means both sketch and photo branches already hold an \textit{explicit} hierarchy, bypassing the need for explicit hierarchy discovery. 
If hierarchies are at all helpful for FG-SBIR, then using explicit hierarchies (with only the {hierarchy discovery module} (Equation~\ref{equ:gumbel}) removed), would provide a weak upper-bound towards sketch-GUI retrieval.
Such explicit hierarchical modelling would just remove the number of nodes at every level by one, until a final d-dimensional vector is obtained.
The result obtained using an \textit{Explicit hierarchy} (71.54\%) is more than that obtained using an \textit{Implicit} one (67.23\%).
This justifies that devising an optimal hierarchy would lead to higher retrieval performance, and shows that our implicit hierarchy discovery module can reach a performance close to its upper bound. As sketches from QMUL-ShoeV2 and QMUL-ChairV2 lack predefined hierarchical structure, implicit hierarchy discovery becomes the only option. 

\begin{figure}
 	\begin{minipage}{0.55\linewidth}
		\centering
		\includegraphics[width=1.0\linewidth, height=0.5\linewidth] {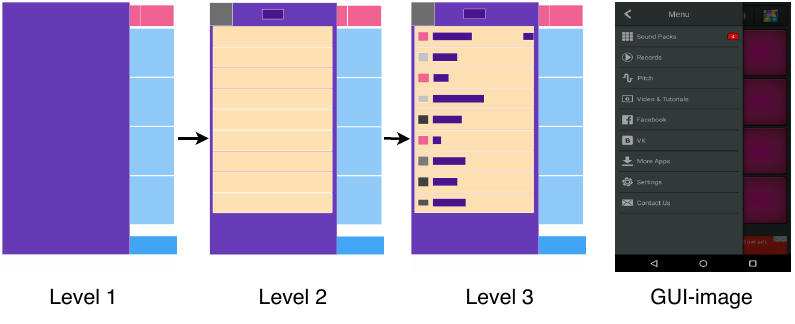} 
		\vspace{-0.3cm}
 		\caption{\textit{Pre-defined} layout-order in a GUI}
		\label{fig:layout}
 	\end{minipage}
 	\begin{minipage}{0.45\linewidth}
 	    \centering
         	\begin{tikzpicture}
                \begin{axis}[
                scale only axis,
                % axis x line=center,
                xlabel=$k\rightarrow$,
                x label style={at={(axis description cs:0.1,0.16)}, anchor=north, font=\footnotesize},
                xmin=0, xmax=5,
                xtick= {1, 2, 3, 4},
                xticklabels={8, 12, 16, 20},
                %xmajorgrids=true,
                %xlabel near ticks,
                %ylabel near ticks,
                y label style={at={(axis description cs:0.24,.5)},anchor=south, font=\footnotesize},
                axis y line* =left,
                ylabel=$\mathrm{Time\;(s)} \rightarrow$,  %\ref{pgfplots:plot1}
                ymin=0, ymax=1.8,
                ytick= {0.3, 0.6, 0.9, 1.2, 1.5},
                yticklabels={},
                %ymajorgrids=true,
                tick label style={font=\scriptsize},
                grid style=dashed,
                legend pos= north west,
                legend cell align={left},
                legend style = {draw=none, font = \scriptsize},
                width=5cm, height=3.5cm
                %area legend
                ]

                \addplot  %dummy
                [ color = red, mark = *, mark size = 2pt, thick, draw=none ]
                coordinates { (1,0.16) (2,0.36) (3,0.67) (4, 1.06)   };
                \addlegendentry{acc@1}
                \addplot 
                [ color = blue, mark = square*, mark size = 2pt, thick ]
                coordinates { (1,0.16) (2,0.36) (3,0.67) (4, 1.06)   };
                \addlegendentry{time}
                
                \addplot [blue, nodes near coords=0.16,
                every node near coord/.style={font=\scriptsize}]
                coordinates {( 0.9, 0.18)};                  
                \addplot [blue, nodes near coords=0.36,
                every node near coord/.style={font=\scriptsize}]
                coordinates {( 2.25, 0.14)};                  
                \addplot [blue, nodes near coords=0.67,
                every node near coord/.style={font=\scriptsize}]
                coordinates {( 3.2, 0.45)};                  
                \addplot [blue, nodes near coords=1.06,
                every node near coord/.style={font=\scriptsize}]
                coordinates {( 3.6, 1.0)};
                \label{pgfplots:plot1}
                \end{axis}
                
                \begin{axis}
                [
                scale only axis,
                xmin=0, xmax=5,
                xtick= {1, 2, 3, 4},
                ylabel=$\mathrm{Top 1 \; Accuracy \; (\%)} \rightarrow$,
                %{Top 1 Accuracy (\%)},  % \ref{pgfplots:plot2}
                ymin=55, ymax=65,
                axis y line*=right,
                y label style={at={(axis description cs:1.2,.5)},anchor=south, font=\footnotesize},
                ytick= {57, 59, 61, 63},
                yticklabels={},
                axis x line=none,
                width=5cm, height=3.5cm
                ]
                \addplot 
                [ color = red, mark = *, mark size = 2pt, thick ]
                coordinates { (1,58.37) (2,60.92) (3,62.45) (4,62.46) };
                
                \addplot [red, nodes near coords=58.37,
                every node near coord/.style={font=\scriptsize}]
                coordinates {( 0.9, 58.67)};                  
                \addplot [red, nodes near coords=60.92,
                every node near coord/.style={font=\scriptsize}]
                coordinates {( 1.9, 61.22)};                  
                \addplot [red, nodes near coords=62.45,
                every node near coord/.style={font=\scriptsize}]
                coordinates {( 2.9, 62.75)};                  
                \addplot [red, nodes near coords=62.46,
                every node near coord/.style={font=\scriptsize}]
                coordinates {( 3.9, 62.76)};
                \label{pgfplots:plot2}
                \end{axis}
            \end{tikzpicture}
        \vspace{-0.7cm}
        \caption{Study on number of regions (k)}
		\label{fig:regions}
 	\end{minipage}
	\vspace{-0.5cm}
\end{figure}

% \vspace{+0.2cm}
\keypoint{Further Analysis:}
\label{sec:further}
(i) Had we skipped cross-modal feature matching (\textbf{w/o Localised-Coattn}) at every hierarchical level, the performance would drop by 10.6\% (4.45\%) for QMUL-ChairV2 (QMUL-ShoeV2) respectively (Table~\ref{tab:ablative}); however it has a lower time cost. On the other hand, eliminating hierarchical modelling (\textbf{w/o Hierarchy}) drops the performance by 7.21\% (3.06\%). This justifies the contribution of both these modules in our method.
(ii) Instead of a distance based ranking in the shared embedding space between query and target photo, one alternative could be to predict their \textit{scalar similarity value} by passing the concatenated features through a two-layer MLP followed by a sigmoid as inspired from \cite{varior2016gated,wang2019camp}. Empirically, performance drops to 53.32\% (32.65\%) justifying the superiority of distance based ranking.
(iii) It has been observed in QMUL datasets that the first few contour strokes are usually long~\cite{yu2016sketch}, constituting majority ($\approx$ 50\%) of the whole sketch (pixel-wise) thus denoting a coarse version of the sketch, while later strokes denote fine-grained details~\cite{yu2016sketch}.
To justify our claim of learning a robust sketch-representation against varying levels of detail, we test our model on sketches where out of all strokes remaining after 50\% pixel-wise sketch completion; (a) half (\textit{Sketch-coarse}) and (b) all (\textit{Sketch-coarse++}); are dropped (Figure~\ref{fig: retrieval}). For SWIRE, we drop inner connected components to do the same. 
{It may be noted that these two settings would yield different, yet similar number of hierarchical levels during training. More specifically, given a sketch-photo pair, the sketch, being typically incomplete, will result in its corresponding hierarchy to reduce to one node sooner compared to the photo, whereas the hierarchical fusion will continue in the photo branch, till it reduces to one node as well.}
{With retrieval being the objective, we achieve relatively stable performance scoring acc.@10 of 87.58\% (77.23\%) and 85.64\% (75.91\%) with \textit{Sketch-coarse} and \textit{Sketch-coarse++} respectively, on QMUL-ChairV2 (QMUL-ShoeV2) datasets. The same for \textit{B-Siamese} however falls to 75.32\% (62.68\%) and 65.31\% (54.32\%). Please refer to Table~\red{1} for comparison against original acc.@10 values. Table~\ref{tab:ablative} shows that even acc.@1 for our method does not drop much in such settings}. Qualitative comparisons are given in Figure~\ref{fig: retrieval}.
{Increasing number of regions (k) chosen for the photo branch (Figure.~\ref{fig:regions}) elevated time cost, while decreasing so chipped at accuracy, proving 16 to be optimal.
}

% As evident from Table~\red{2} performance stays relatively stable for our method, especially when compared with the \textit{B-Siamese} alternative. Qualitative comparisons are given in Figure~\ref{fig: retrieval}.
% (iv) The advantage of paired embedding comes at an  additional time-cost as compared to independent embedding methods. While B-Siamese takes 0.14 (0.37) milliseconds per query in the evaluation setup, our model takes 55.06 (138.37) milliseconds with respect to QMUL-ChairV2 (ShoeV2) datasets. Desite significant higher, this is still managable for practical applicationsThe high time cost is due to an additional looping over the entire gallery images per query for cross-modal feature learning.

\begin{table}
% 	\begin{minipage}{0.5\linewidth}
	\centering
 	\caption{Ablative Study (acc.@1)}
 	\label{tab:ablative}
  	\resizebox{0.6\columnwidth}{!}{
	\begin{tabular}{lccc}
		\toprule
		{Methods} & {Chair-V2} & {Shoe-V2} & {SWIRE} \\ 
        \midrule
        Explicit Hierarchy          & -  & -   & \bf 71.54 \\ 
        w/o Localised-Coattn        & 51.85 & 31.82  & 60.32 \\ 
        w/o Hierarchy               & 55.24    & 33.21   & 65.48 \\
        Sketch-coarse    & 47.64  & 31.83  & 51.26 \\ 
        Sketch-coarse++  & 42.33  & 24.11  & 45.33  \\
        \hline
        Proposed                   & \bf 62.45  & \bf 36.27 &  67.23 \\
		\bottomrule
	\end{tabular}
 	} 
% 	\end{minipage}%\hfill
\end{table}           

\vspace{-0.5cm}
\section{Conclusion}
\label{sec:conclusion}
\vspace{-0.25cm}
In this paper, we studied a further intrinsic trait of sketches -- that they are hierarchical in nature. The benefit of modelling and utilising hierarchies is demonstrated for the problem of {FG-SBIR}. Our model learns to discover a hierarchy that is implicit to a sketch. This ensures that no matter what level of detail an user sketches to, the model shall be able to retrieve the same/similar image fairly accurately. Unlike earlier approaches, our network enforces a cross-modal co-attention between a sketch and a photo, so as to arrive at a better cross-modal embedding for retrieval. Extensive experiments show our model to outperform most existing approaches by a significant margin, on both image and GUI retrieval.
%Ablative studies justify our design choices as well. fine-grained sketch-base image retrieval

%-------------------------------------------------------------------------

%------------------------------------------------------------------------

\bibliography{egbib}
\end{document}